%% file: 2023_paper_giacobrizi_ca2lib.tex
\documentclass[letterpaper, 10 pt, conference]{ieeeconf}  
\IEEEoverridecommandlockouts                              
\usepackage{graphics}    
\usepackage{times}       
\usepackage{amsmath}     
\usepackage{amssymb}     
\usepackage{graphicx}
\usepackage{algorithm}
\usepackage{glossaries}
\usepackage[noend]{algpseudocode}
\usepackage{booktabs}
\usepackage{makecell}
\usepackage[colorlinks=true,linkcolor=black,citecolor=black,urlcolor=blue]{hyperref} 
\usepackage[font=footnotesize]{caption}
\usepackage[font=footnotesize]{subcaption}
\usepackage{flushend}
\usepackage{footmisc}

\usepackage{balance}

\def\figref#1{Fig.~\ref{#1}}
\def\tabref#1{Tab.~\ref{#1}}
\def\eqref#1{Eq.~(\ref{#1})}

\newcommand\etal{\emph{et al.}}

\def\ie{{i.e.}}
\def\eg{{e.g.}}
\def\etal{\emph{et al.}}
\def\lidar{LiDAR}
\def\lidars{LiDARs}

\def\rgb{RGB}

\input{notation}

\newacronym{slam}{SLAM}{Simultaneous Localization and Mapping}
\newacronym{sfm}{SfM}{Structure from Motion}
\newacronym{pgo}{PGO}{Pose-Graph Optimization}
\newacronym{vpr}{VPR}{Visual Place Recognition}
\newacronym{sgd}{SGD}{Stochastic Gradient Descent}
\newacronym{ils}{ILS}{Iterative Least-Squares}
\newacronym{icp}{ICP}{Iterative Corresponding Point}
\newacronym{gn}{GN}{Gauss-Newton}
\newacronym{lm}{LM}{Levenberg-Marquardt}
\newacronym{pcg}{PCG}{Preconditioned Conjugate Gradient}
\newacronym{map}{MAP}{Maximum-A-Posteriori}
\newacronym{gf}{GF}{Gaussian Filters}
\newacronym{pf}{PF}{Particle Filters}
\newacronym{sdp}{SDP}{Semi-Definite Programming}
\newacronym{bst}{BST}{Binary Search Tree}
\newacronym{ndt}{NDT}{Normal Distributed Transform}
\newacronym{ba}{BA}{Bundle Adjustment}
\newacronym{pnp}{PnP}{Perspective-n-Points}

\def\figref#1{Fig.~\ref{#1}}
\def\tabref#1{Tab.~\ref{#1}}
\def\eqref#1{Eq.~(\ref{#1})}

\usepackage{lipsum}

\title{\LARGE \bf \textit{Ca$^2$Lib}: Simple and Accurate \lidar-\rgb~Calibration \\using Small Common Markers}

\author{
	\large
	Emanuele Giacomini* \quad\quad
	Leonardo Brizi* \quad\quad
	Luca Di Giammarino \quad\quad
	Omar Salem\\\\
	\centering
	Patrizio Perugini \quad\quad
	Giorgio Grisetti
\thanks{All authors are with the Department of Computer, Control, and Management Engineering ``Antonio Ruberti", Sapienza University of Rome, Italy,
Email:\,\,{\tt\footnotesize{\{giacomini, brizi, digiammarino,
		salem, grisetti\}@diag.uniroma1.it.}}}%
  \thanks{This work has been supported by PNRR MUR project PE0000013-FAIR}
  \thanks{* The authors contributed equally.}
}

\begin{document}
\maketitle
\thispagestyle{empty}
\pagestyle{empty}

\begin{abstract}
  %

  In many fields of robotics, knowing the relative position and orientation between two sensors is a mandatory precondition to operate with multiple sensing modalities.
  In this context, the pair \lidar-\rgb~cameras offer complementary features:
  \lidars~yield sparse high quality range measurements, while \rgb~cameras provide a dense color measurement of the environment. 

 Existing techniques often rely either on complex calibration targets that are expensive to obtain, or extracted virtual correspondences that can hinder the estimate's accuracy.
 
In this paper we address the problem of \lidar-\rgb~calibration using typical calibration patterns (\ie~A3 chessboard) with minimal human intervention.
  Our approach exploits the planarity of the target to find correspondences between the sensors measurements, leading to features that are robust to \lidar~noise. 
  Moreover, we estimate a solution by solving a joint non-linear optimization problem.

 
	We validated our approach by carrying on quantitative and comparative experiments with other state-of-the-art approaches. Our results show that our simple schema performs on par or better than other approches using complex calibration targets.
	Finally, we release an open-source C++ implementation at \url{https://github.com/srrg-sapienza/ca2lib}
\end{abstract}

\section{Introduction}
\label{sec:intro}

The ability to fuse readings from heterogeneous sensors is often beneficial in many robotics and perception applications. In particular, \lidar~and \rgb~sensors exhibit a strong compatibility: the former being able to capture high precision sparse range readings while the latter measure dense color intensity measurements.



These properties makes integration between the two sensors suited for the task of \emph{depth estimation}.
Historically, stereo based solutions leverage the known relative offset between two cameras, along with concepts from epipolar geometry to estimate a depth value for every pixel in a image. Albeit its popularity, due to its optical nature, these approaches suffer in texture-less regions and in areas where the depth exceed a maximum value determined by the baseline of the stereo. While the texture-less problem has been partially solved by the usage of active stereo sensors (\ie~Realsense D435, Kinect), the maximum range still poses a challenge.
On the contrary, \lidars~operates using Time of Flight (TOF) principle, which is proven to be robust at detecting range measurements on non-reflective surfaces with accuracy even at high distances.

These considerations lead the community to investigate the problem of \emph{depth-completion}, namely estimating a dense depth image by superimposing an accurate sparse depth measurement along an intensity \rgb~image.
Multiple publicly available datasets like KITTI and VOID \cite{uhrig2017}\cite{wong2020} allowed the community to tackle this problem either by fully leveraging the sparse depth measurement (unguided) or by fusing \rgb~features (guided).
Furthermore, in the field of 3D reconstruction, recent findings show that coupling the two sensors may lead to a more robust and accurate trajectory estimate \cite{di2023photometric}.

Besides, to accomplish any of these tasks, one would require to know the relative offset between the two sensors.

This work aims at solving the task of \lidar-\rgb~calibration, namely, estimating the relative offset (extrinsic parameters) between the two sensors, using their raw measurements.

\begin{figure}[t]
	\vspace{0.3cm}
	\centering
	\includegraphics[width=0.99\linewidth]{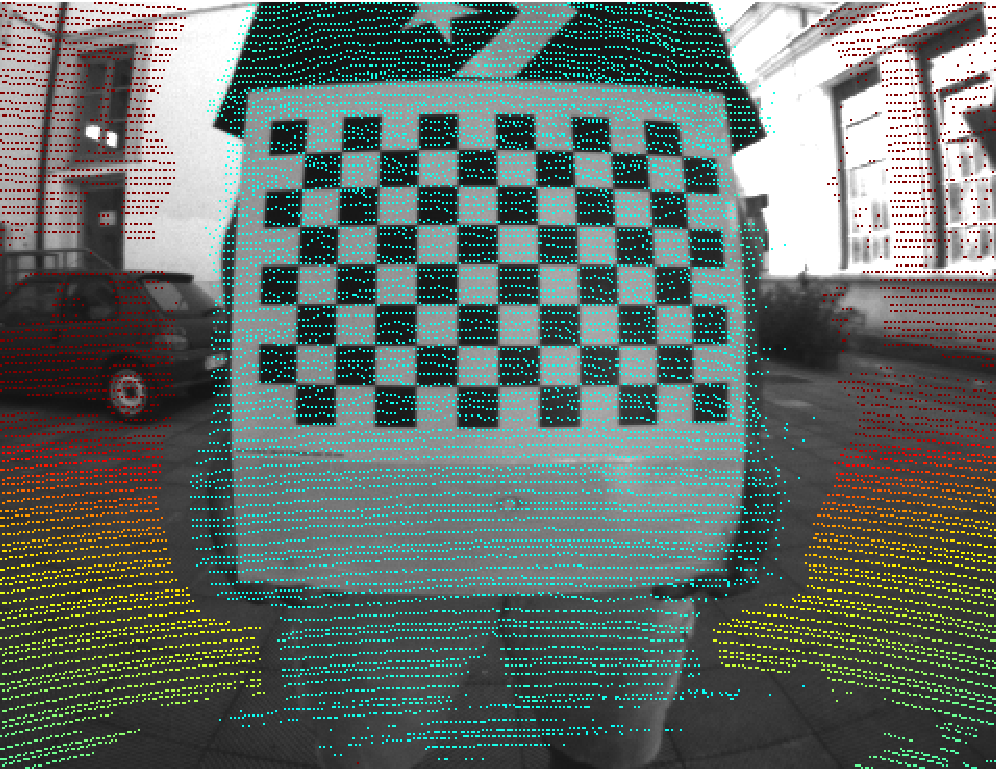}
	\caption{Reprojection of a \lidar~point cloud on a fisheye \rgb~camera rigidly attached to the former. The offset between the sensors leads to shadows on parts of the image.}
	\label{fig:motivation}
\end{figure}

The core idea behind this multi-modal calibration is to find spatial correspondences between the sensors measurements. Most approaches rely on one or more calibration patterns to establish common features between the sensors. These patterns are often complex or expensive to produce \cite{beltran2022}.
The main contribution of this paper is a versatile calibration toolbox that allows to estimate the extrinsic parameters between \lidar~and \rgb~with minimal user intervention, using a simple calibration checkerboard target.  We leverage a joint non-linear formulation of the problem to achieve high accuracy results even with a minimum of three measurements. The requirement for our method is to use a calibration pattern (\eg~Checkerboard, ChAruCO \cite{garrido2014}) that must be observed by both sensors during the acquisition. We exploit the planarity of the target to find a common observation used to estimate the extrinsic parameters. Moreover, we release an open-source implementation of our toolbox.

\section{Related Work}
\label{sec:related}

This paper's section delves into \lidar-\rgb~calibration and explores the two main classes of approaches: \emph{target-based} and \emph{target-less}. As the name suggests, target-based approaches require the user to place artificial markers that both the camera and \lidar~can easily detect. This contrasts with target-less methods that free the use from this task. The core idea of calibration is common in the two classes of approaches: computing common features between heterogeneous measurements and estimating the transformation that minimizes the distance between corresponding features.

First, an overview of \emph{target-less} approaches is presented:
Pandey \etal~presents an automatic data-driven approach based upon the maximization of mutual information between the sensor-measured surface intensities \cite{pandey2015}. The authors exploits the correlation coefficient for the reflectivity and intensity values of many scan-image pairs using different calibration parameters. However, shadows of objects or colored surfaces that completely absorb infrared lights might result in weaker correlation between scan-image pairs.
Yoon \etal~proposes a calibration method using region-based object pose estimation. Objects are segmented in both measurements, then a 3D mesh is generated from the \lidar~measurements, while images are used to reconstruct the scene using \gls{sfm}. The two models are then registered together to acquire an initial guess on the relative pose. The final solution is obtained iteratively by finding correspondences between the reconstructed objects from both measurements \cite{yoon2021}.
In recent years, the development of learning based methods have also spanned in this field: Lv \etal~ proposes a real-time self-calibration network that predict the extrinsic parameters by constructing a cost volume between \rgb~and \lidar~features \cite{lv2021}, while Sun \etal~first estimates an initial guess by solving an hand-eye calibration method \cite{sun2022}. Moreover, the guess is fine-tuned by segmenting the image-cloud pair and by aligning the distances between centroids.
The advantage of target-less method is that they can be used without preparing the environment. This comes at a cost of a lower accuracy and robustness when compared to their target-based counterpart.

\emph{Target-based} methods estimate the relative pose using an observed known structure. Given the difference of resolution for the two sensors, it is high unlikely that correspondences within the measurements can be established directly. For this reason, \emph{point-to-point} methods tends to process \lidar~measurements to implicitly obtain virtual points\footnote{Points that are not explicitly detected, but estimated from the \lidar~measurement.} easily detectable from an \rgb~sensor. For instance, Park \etal~utilizes a specially designed polygonal planar calibration board with known lengths of adjacent sides \cite{park2014}. By estimating the 3D corresponding points from the \lidar, vertices of the board can be determined as the meeting points of two projected sides. The vertices, along with the corresponding points detected from the color image, are used for calibration. Pusztai \etal~introduces a methodology that utilizes cubic objects with predetermined side lengths \cite{pusztai2017, pusztai2018}. The corners of the cubes are estimated by initially detecting each side of the box and subsequently determining their intersection points. Furthermore, the corners along with their corresponding \rgb~image are employed to calibrate the system by solving \gls{icp}. \emph{Zhou}\etal~ proposes a single-shot calibration method requiring a checkerboard \cite{zhou2018}. The target is detected both in the RGB image, and \lidar~measurement, using RANSAC\cite{fischler1981} for the latter. Furthermore, the four edges of the checkerboard are estimated and aligned to compute the relative offset between the two sensors. Tóth \etal~introduces a fully automatic calibration technique that leverages the utilization of spheres, enabling accurate detection in both point clouds and camera images \cite{toth2020}. Upon successful detection, the algorithm aligns the set of sphere centers using SVD. Beltrán \etal~presents a methodology that utilizes a custom calibration target equipped with 4 holes and AruCO markers specifically designed for monocular detection \cite{beltran2022}. The methodology employs a set of techniques for each sensor to estimate the center points of the holes. Subsequently, the relative offset between sensors are determined by aligning the set of centers obtained from each sensor, Li \etal~adopt a similar approach while using a checkerboard with 4 holes \cite{li2022}. Fan \etal~propose a two-stage calibration method using an auxiliary device with distinctive geometric features \cite{fan2023}. The method extracts lines from images and \lidar~point clouds, providing an initial estimation of the external parameters. Nonlinear optimization is then applied to refine these parameters. In the work of Singandhupe \etal, the authors first extract planar information from \rgb~and \lidar~measurements, then, two grid of points are extracted from the computed planar patches and aligned using a customized \gls{icp} algorithm \cite{singandhupe2022}.

Albeit these approaches provides relatively accurate results with few measurements, care should be taken during the estimation of virtual correspondences, as they can cause significant errors in the estimation step. Moreover these custom targets often requires precise construction or expensive manufacturing.

Another group of approaches does not directly solve the calibration problem using point-to-point correspondences, but rather exploit the planarity of the target to reduce the feasible set of solutions using \emph{plane-to-plane} constraints. Mirzaei \etal~addresses the challenge of accurate initial estimates by dividing the problem into two sub-problems and analytically solving each to obtain precise initial estimates \cite{mirzaei2012}. The authors then refine these estimates through iterative minimization. They also discuss the identifiability conditions for accurate parameter estimation.
Finally, a method similar to our proposal, Kim \etal~ combine observed normals to first estimate the relative orientation with SVD and then iteratively estimates an initial guess of the relative translation by minimizing the pairwise planar distances between measurements \cite{kim2020}. Finally, the translation is refined using a non-linear optimization problem using \gls{lm}. Despite its simplicity, this method decouples the estimation of orientation and translation, thus leading to potential losses in accuracy while also increasing the number of required measurements.

Compared with the state of the art, we propose:
\begin{itemize}
	\item a formulation for joint nonlinear optimization that couples relative rotation and translation using a plane-to-plane metric;
	\item an extensible framework that decouples the optimization from target detection. Currently supports Checkerboard and \emph{ChARuCO} patterns of typical A3-A4 sizes, easily obtainable from commercial printers; 
	\item the possibility to handle different camera models and distortion;
	\item an open-source C++ implementation.
\end{itemize}

\section{Our Approach}
\label{sec:main}
In this section, we will provide a detailed and comprehensive description of our method.
First we describe the preliminaries required to understand our approach, then every component of the pipeline is described, following the procedure from the acquisition of the measurements up to the computation of the relative poses between the two sensors (extrinsic parameter).

\textbf{Plane Representation:}
Let $\pi = (\hat{\bn}, d)$ be a 3D plane, where $\hat{\bn} \in \mathbb{S}^2$ represents the unit vector orthogonal to the plane and $d \in \bbR$ is the shortest distance of the plane respect to the origin. Applying a transform $\bX \in \bbSE(3)$ to a plane $\pi$ yields new coefficients $\pi'$:
 as follows:
\begin{equation}
\bX\pi = \left\{ 
\begin{array}{ll}
\hat{\bn}' = &\bR \bn\\
d' = &d + (\bR \bn)^t \bt			
\end{array}
\right .
\end{equation}
Here $\bX = \left<\bR; \bt\right>$ is represented by a rotation matrix $\bR\in \bbSO(3)$, and the translation vector $\bt\in \bbR^3$.

If the transformation is modified by a small local perturbation $\bDeltaX=(\bDeltaR \vert \bDeltat)$ then we can rewrite:
\begin{equation}
(\bX \boxplus \bDelta\bX)\pi = 
\left\{
\begin{array}{ll}
\tilde{\bn} = &\bDeltaR \bR \bn\\
\tilde{d} = &d' + \bn^t\bR^t\bDeltaR^t \bDeltat
\end{array}
\right.
\end{equation}
Deriving the result with respect to $\bDelta\bX$ leads to the following Jacobian:
\begin{equation}
\frac{\partial (\bX \boxplus \bDelta\bX)\pi}{\partial \bDelta\bX} = 
\begin{bmatrix}
&0_{3 \times 3} &-\skew{\bR \bn}\\
&\bn^t\bR^t &0_{1 \times 3}
\end{bmatrix}_{4 \times 6}
\end{equation}

The distance between two planes depends both on the difference between their normals and the signed distance of the planes from the origin. These quantities can be captured by a 4D error vector $e_p$ expressing the \emph{plane-to-plane} error metric:
\begin{align}
\bp(\pi_k) &= -\bn_k d_k \\
e_p(\pi_i, \pi_j) &= 
\begin{bmatrix}
	\bn_i^t (\bp(\pi_i) - \bp(\pi_j))\\
	\bn_j -\bn_i
\end{bmatrix}.
\end{align}
Here $\bp(\pi_k)$ is the point on the plane closest to the origin of the reference system, and it is obtained by taking a point along the normal direction $\bn$ at distance $d$.

\textbf{Pinhole Model (\rgb):} Let $\bp$ be a point expressed in camera frame and $\bK$ be t camera matrix. Assuming any lens distortion effect have been previously corrected, then the projection on the image plane of $\bp$ is computed as

\begin{align}
\pi_{\mathrm{c}}(\bp) &= \phi(\bK\bp)\\
\bK &= \begin{bmatrix}
f_x &0 &c_x\\ 0 &f_y &c_y\\0&0&1
\end{bmatrix}\\
\phi(v) &= \frac{1}{v_z}\begin{bmatrix}
	v_x\\v_y
\end{bmatrix}
\end{align}
where $\phi(v)$ represents the homogeneous division and $\pi_{\mathrm{c}}(\bp)$ the pinhole projection function. For simplicity, we detail only the pinhole camera projection, however the same principle applies for more complex camera models. 

\textbf{Projection by ID (\lidar):} Let $\bp$ be a point detected by the \lidar~and expressed in its frame. Its projection is computed as:
\begin{align}
\pi_{\mathrm{l}}(\bp) &= \bA\psi(\bp)\\
\bA &= 
	\begin{bmatrix}
		f_x &0 &c_x\\
		0 & 1 &0
	\end{bmatrix}\\
\psi(v) &= 
	\begin{bmatrix}
		\atantwo(v_y, v_x)\\
		\textit{ring}(v)\\
		1
	\end{bmatrix}	
\end{align}
where $f_x$ represent the azimuth resolution of the \lidar, while $c_x$ denotes the offset in pixels. The $\textit{ring}(v)$ function is either obtained directly from the \lidar~sensor, which augment every measured point with a number that represents the beam that detected it or, assuming the cloud is ordered, by dividing the point index by the horizontal resolution of the sensor. Compared with the classical spherical projection, the projection by ID does not preserve the geometric consistency of the scene. Still, it provides an image with no holes, which is preferred for computer-vision applications.

First, we process the incoming raw \lidar~and \rgb~measurements to acquire planar information. Assuming the scene to remain static throughout the acquisition of a single joint measurement, the \lidar~cloud is embedded in an image using the projection by ID. Moreover, the system awaits the user interaction to guess the position of the calibration target on the \lidar~image.

A parametric circular patch around the user's selection is used to estimate a plane using RANSAC and, concurrently, the calibration target detection is attempted on the \rgb~image. Once the target is detected, the \rgb~plane is computed by solving the \gls{icp}. If the user is satisfied with both \lidar~and \rgb~planes, they are stored for processing.

Whereas a straightforward rank analysis of the Jacobians reveals that just  3 measurments are sufficient to constrain a solution, it is well known from the estimation theory that the accuracy grows with the number of measurements.

Once the set of measurements are acquired, we jointly estimate the relative orientation and translation of the \lidar~ with respect to the \rgb~sensor $\bX \in \bbSE(3)$ by solving the following nonlinear minimization problem:
\begin{equation}
\label{eq:ls}
\bX = \argmin_{\bX \in \bbSE(3)} \sum_{i\in\bZ} \lVert \underbrace{\bX\pi^i_\mathrm{l} -\pi^i_\mathrm{c}}_{e_p} \rVert^2
\end{equation}
where $e_p$ represent the plane-to-plane error.

During acquisition, it may happen that the user accept one or more wrongly estimated measurements. Due to the quadratic nature of the error terms, these \emph{outliers} are often over-accounted, resulting in wrong estimations. To account for this factor, as described in \cite{grisetti2020}, we employ an Huber \emph{M-estimator} $\rho(\cdot)$ that treats differently measurements based on their error. We rewrite \eqref{eq:ls} as follows:
\begin{equation}
\label{eq:irls}
\bX = \argmin_{\bX \in \bbSE(3)} \sum_{i\in\bZ} \rho(\lVert \bX\pi^i_\mathrm{l} -\pi^i_\mathrm{c} \rVert).
\end{equation}
To resolve \eqref{eq:irls} we employ the \gls{gn} algorithm implemented in the \texttt{srrg2\_solver}\cite{grisetti2020}.

\section{Experimental Evaluation}
\label{sec:exp}
\begin{table*}
	\centering
	\begin{tabular}{lrrrrrr}
		\toprule
		{} &  \multicolumn{2}{c}{\makecell{$\sigma_{\mathrm{l}}=0$\\$\sigma_{\mathrm{c}}=0$}} & \multicolumn{2}{c}{\makecell{$\sigma_{\mathrm{l}}=8e^{-3}$\\$\sigma_{\mathrm{c}}=7e^{-3}$}} & \multicolumn{2}{c}{\makecell{$\sigma_{\mathrm{l}}=16e^{-3}$\\$\sigma_{\mathrm{c}}=14e^{-3}$}} \\
		{} &    mean &    stdev &    mean &   stdev &    mean &    stdev \\
		\midrule
		\textbf{No. Measurements} &         &          &         &         &         &          \\
		\midrule
		\textbf{3               } &  41.761 &  104.362 &  20.790 &  25.124 &  57.849 &  112.365 \\
		\textbf{4               } &  10.872 &   17.941 &  12.206 &  12.363 &  14.940 &   11.681 \\
		\textbf{5               } &   6.492 &    7.997 &   8.350 &   9.076 &   9.115 &    5.675 \\
		\textbf{10              } &   4.591 &    3.458 &   5.759 &   4.974 &   5.849 &    1.989 \\
		\textbf{20              } &   2.575 &    1.981 &   3.646 &   2.564 &   4.123 &    1.139 \\
		\textbf{30              } &   2.673 &    1.263 &   2.867 &   1.659 &   3.735 &    0.878 \\
		\textbf{39              } &   2.091 &    0.883 &   2.666 &   1.206 &   3.261 &    0.413 \\
		\bottomrule
	\end{tabular}
	\caption{Average translation error in millimeters with different noise levels and number of measurements.}
	\label{tab:exp-synt}
\end{table*}

In this section, we describe the experiments we conducted to establish the quality of our calibration toolbox.
We perform quantitative experiments in the simulated environment provided by \cite{beltran2022} to compare our estimates with groundtruth while we also conduct qualitative and quantitive experiments on real scenarios using our acquisition system. We directly compare our results with \cite{kim2020}, as it is the work which is closest to ours. In addition, we compare to \cite{beltran2022} that produce accurate results relying on a very complex target (CNC printed).

\subsection{Synthetic Case}
We conducted experiments on \emph{Gazebo} simulator \cite{koenig2004} to evaluate the accuracy and robustness of our approach, injecting different noise figures to the sensor measurements. We also experiment how the number of observations affect the final results.
The setup of the scene includes a Velodyne HDL-64 \lidar, a BlackFly-S \rgb~sensor and a $6\times 8$ checkerboard target with corner size of $0.2$ meters. We randomly generate and acquire $53$ valid\footnote{A valid measurement is one for which both \lidar~and \rgb~sensor are able to detect the target} measurements.

To quantify the impact of the number of measurements on the accuracy of our approach, we run the calibration procedure with an increasing number of measurements $w_s=[3\dots39]$ and at three different \lidar~noise levels $\sigma_{\mathrm{l}}$ ($0$ mm, $7$ mm and $14$ mm). For every $w_s$, we sample $40$ sets of measurements. 

From \tabref{tab:exp-synt}, we observe a steady decrease of error for every noise level, reaching an average of $2.6$mm translation error in the intermediate noise case. In case of $3$ measurements the high uncertainty is due to the potentially poorly conditioned system when using planes that have similar normals.
Nonetheless, we compare our best result with $3$ measurements against the best results of the methods presented in \cite{beltran2022} and \cite{kim2020}. \tabref{tab:comparison-synt} shows the results.

\begin{table}[h]
	\centering
	\begin{tabular}{lcc}
		\toprule
		Method & $e_t$ (cm)    & $e_r$ ($10^{-2}$ rad) \\
		\midrule
		\emph{Beltrán} \etal\cite{beltran2022}      & 0.82          & 0.50                  \\
		\emph{Kim} \etal\cite{kim2020}      & 10.2          & 129.56                \\
		Ours   & \textbf{0.11} & \textbf{0.25}         \\ \hline
	\end{tabular}
	\caption{Best solutions obtained by calibration using $N=3$ measurements.}
	\label{tab:comparison-synt}
\end{table}	

\subsection{Real Case}



\begin{figure}[t]
	\centering
	\includegraphics[width=0.99\columnwidth]{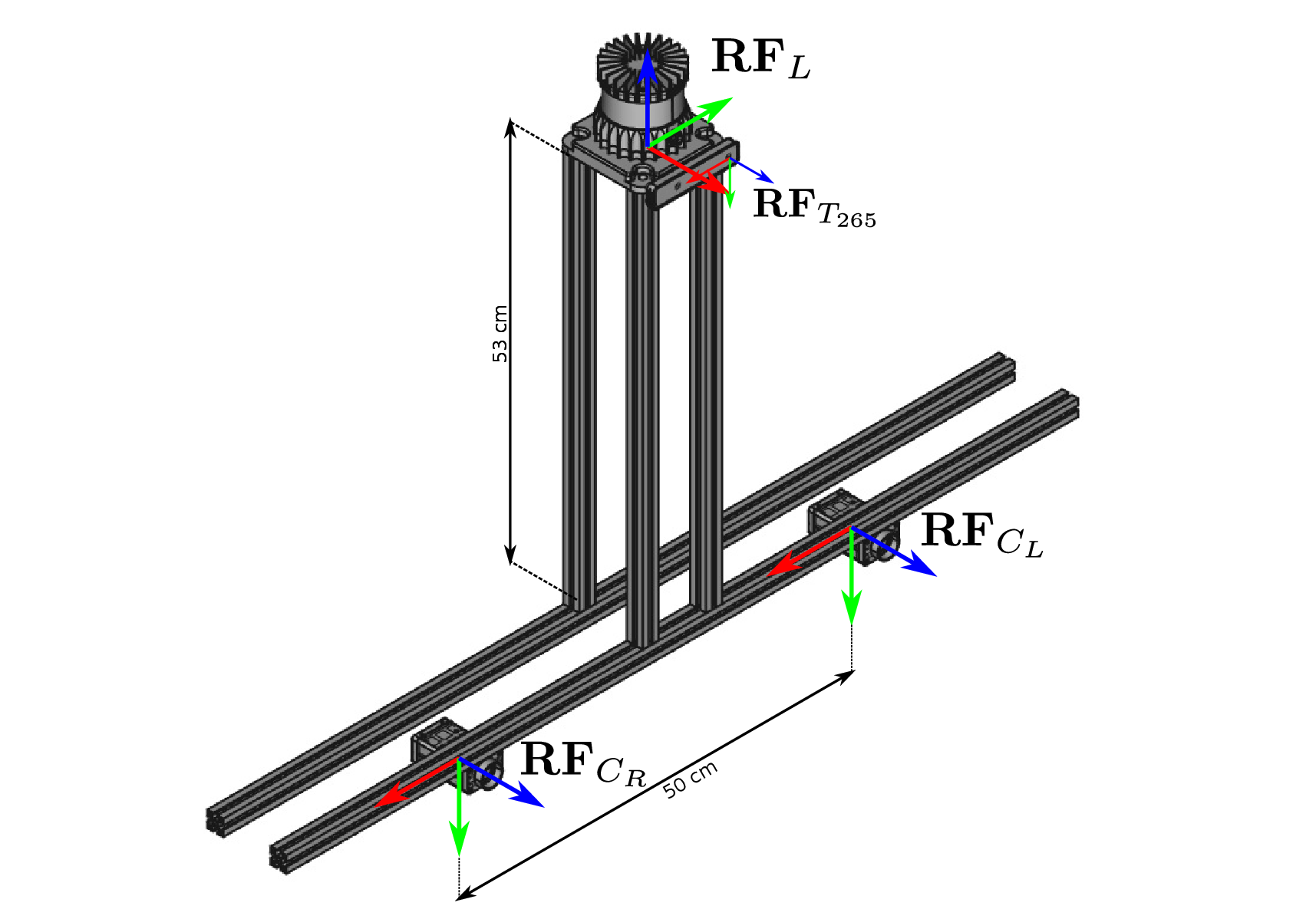}
	\caption{Acquisition system used for the Real Case experiments.}
	\label{fig:sbdg}
\end{figure}

In this section, we describe the experiments conducted on real measurements. We perform a quantitative test on our acquisition system shown in \figref{fig:sbdg} that is equipped with a Ouster OS0-128 \lidar~with a resolution of $128 \times 1024$, a RealSense T-265 stereo camera and two Manta-G145 \rgb~cameras arranged in a wide horizontal stereo configuration.

Since no groundtruth information is available, we take advantage of the stereo extrinsics to provide an estimate of the calibration error. The offset between multiple camera is measured using optical calibration procedures which typically reach subpixel precision.
 
In the first experiment, we consider the \lidar~and the Realsense T-265 sensor which provides factory calibrated intrinsic/extrinsic parameters for both cameras. The task of the experiment is to demonstrate the accuracy of the calibrator in real case scenarios and to understand how the number of measurements considered affects the quality of the solution.

\begin{figure}[h]
	\centering
	\includegraphics[width=0.99\columnwidth]{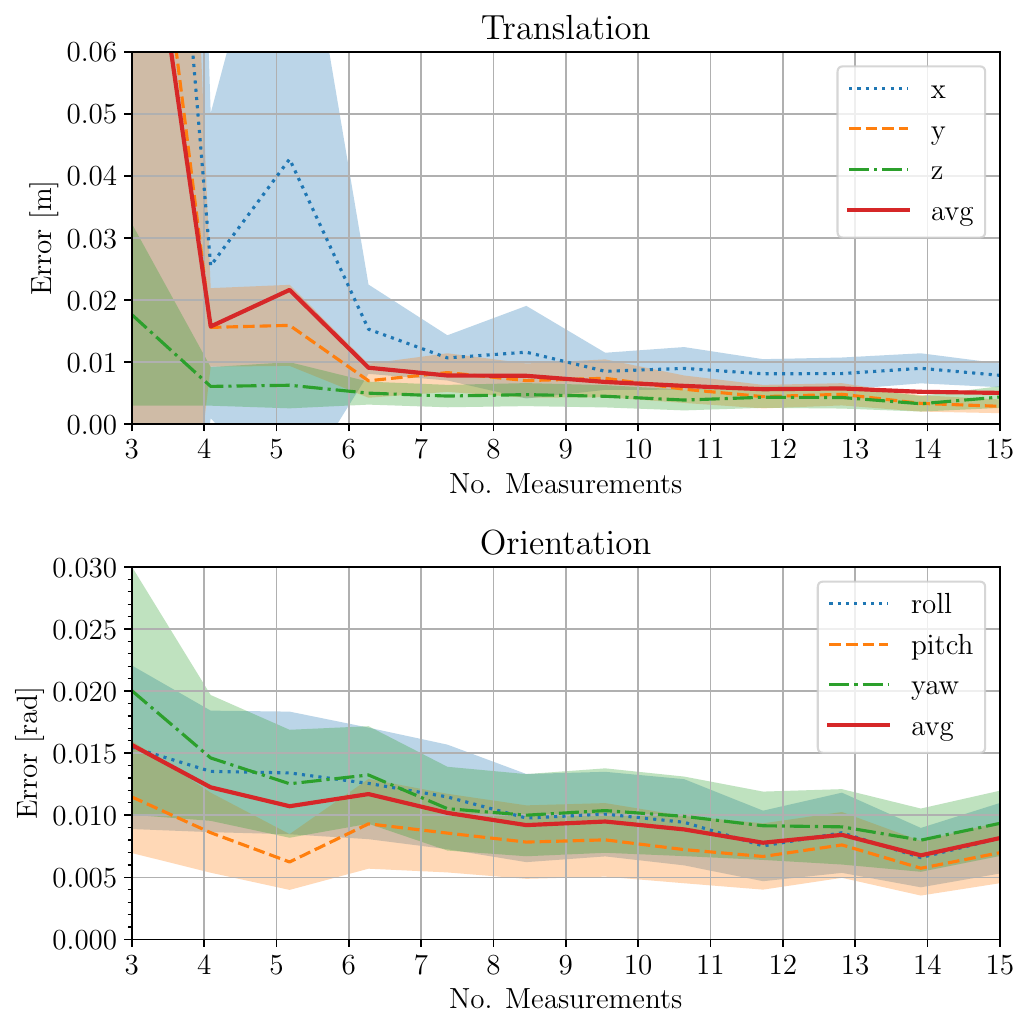}
	\caption{Average camera-wise calibration error in the \lidar-T265 case.}
	\label{fig:t265_stereo}
\end{figure}

As for the synthetic case, we first acquire a set of $17$ cloud-image \lidar~\rgb~measurements for both cameras. Moreover we perform $40$ calibrations with $w_s$ randomly selected measurements with $w_s\in\{3, 15\}$. Finally, for every $w_s$, we combine the computed extrinsics for both cameras to obtain an estimate stereo transform. Assuming approximately symmetrical errors in the two cameras, \figref{fig:t265_stereo} shows the results of this experiment. We were able to obtain at best an average error of $7.1$ mm in translation and $0.01$ rads in orientation.

\begin{figure}[h]
	\centering
	\includegraphics[width=0.99\columnwidth]{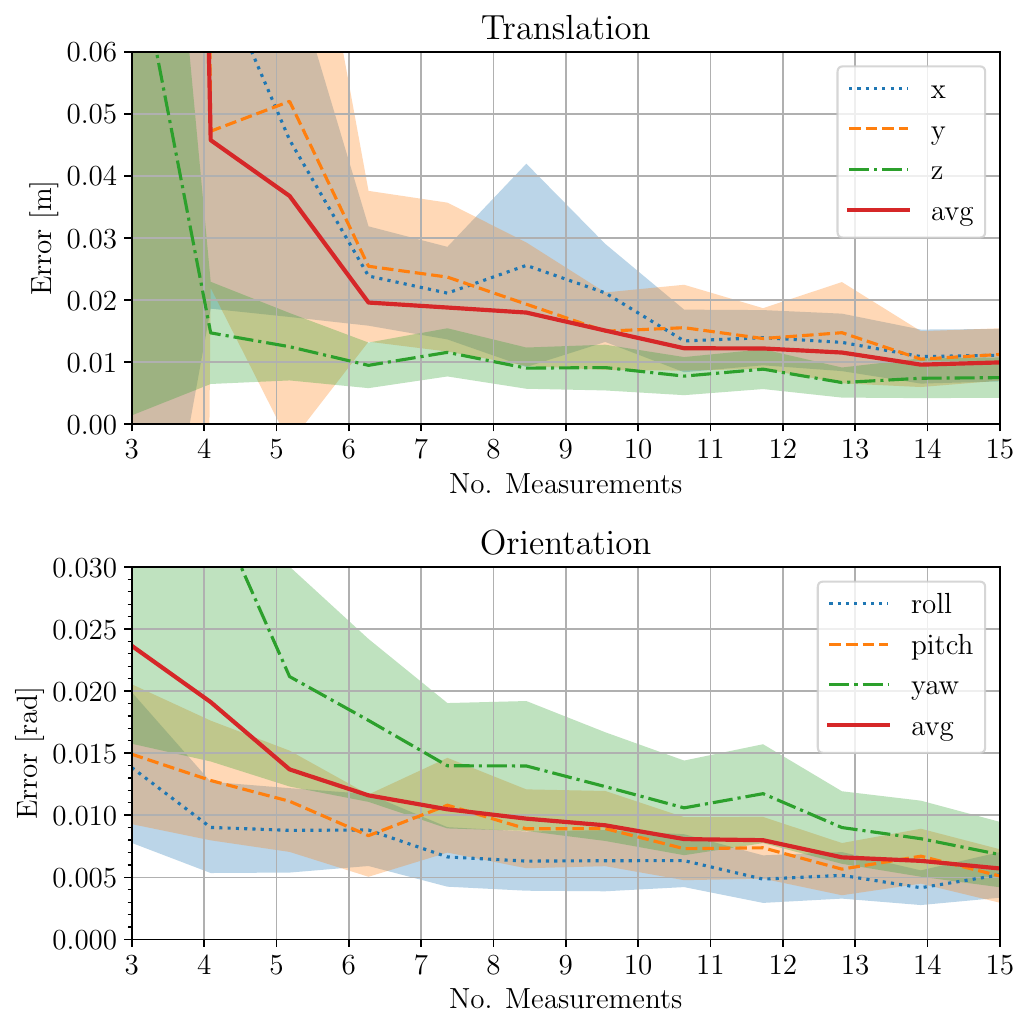}
	\caption{Average camera-wise calibration error in the \lidar-Manta case.}
	\label{fig:manta_stereo}
\end{figure}

The second experiment is conducted using the wide stereo setup for which we also calibrate the intrinsics and extrinsics of the cameras, providing expected results in a typical scenario. The acquisition procedure is the same as in the first experiment and \figref{fig:manta_stereo} shows the experimental result, where we obtain the best solution with $4.6$ mm and $0.002$ rads in orientation.

\begin{figure*}
\centering
\begin{subfigure}{0.49\textwidth}
	\centering
	\includegraphics[width=\textwidth]{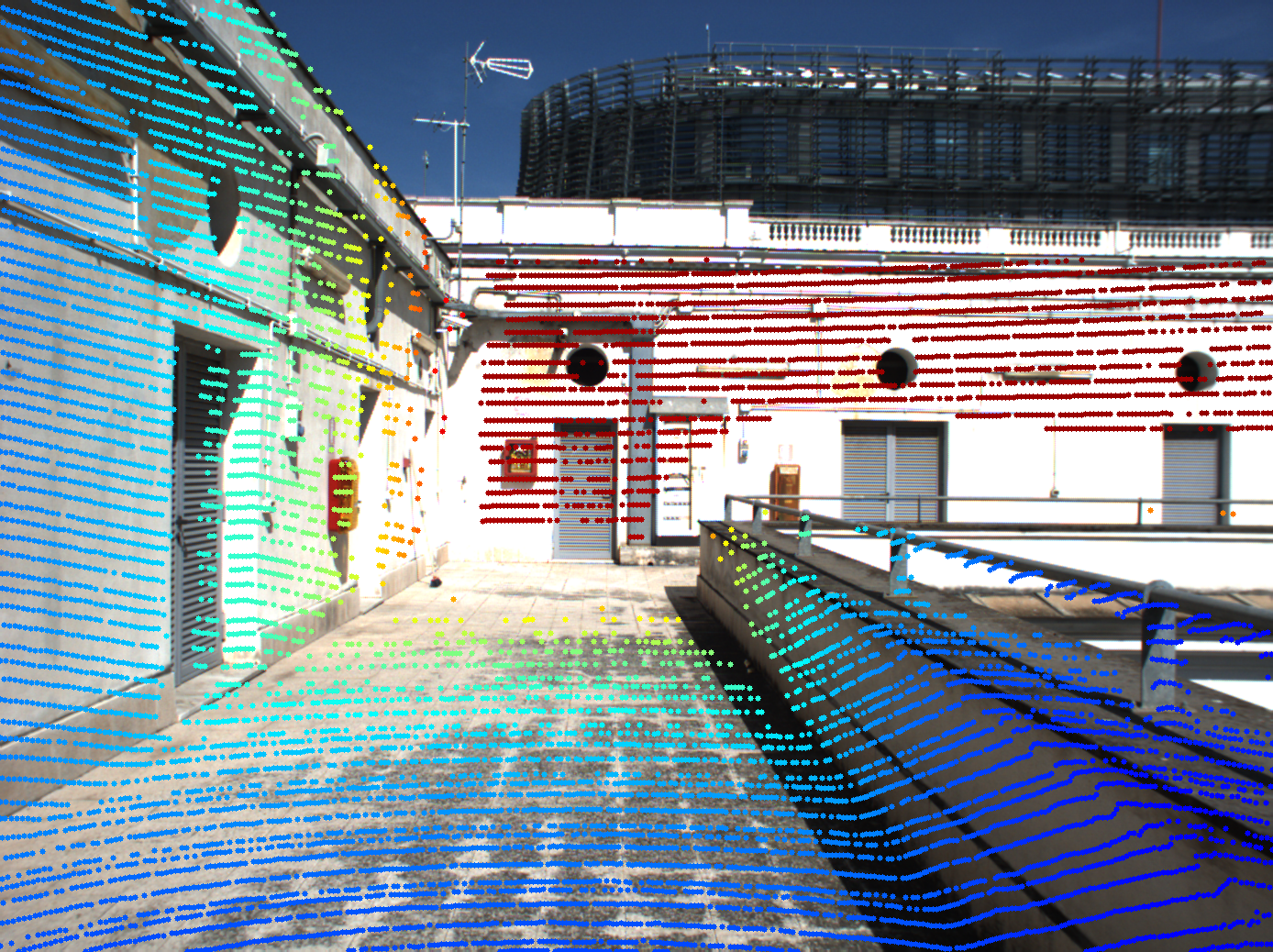}
\end{subfigure}
\begin{subfigure}{0.49\textwidth}
	\centering
	\includegraphics[width=\textwidth]{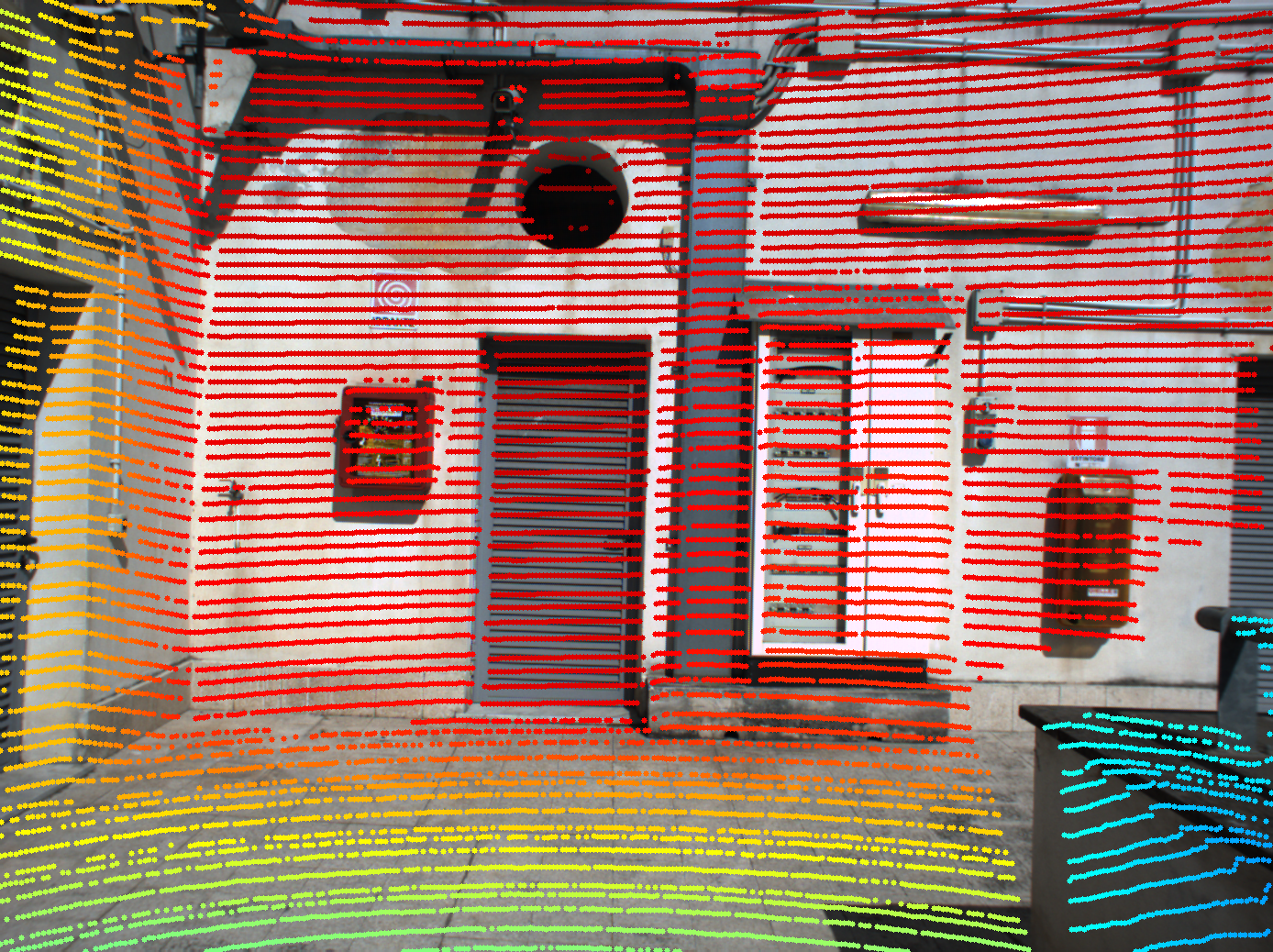}
\end{subfigure}
\begin{subfigure}[b]{0.49\textwidth}
	\centering
	\includegraphics[width=\textwidth]{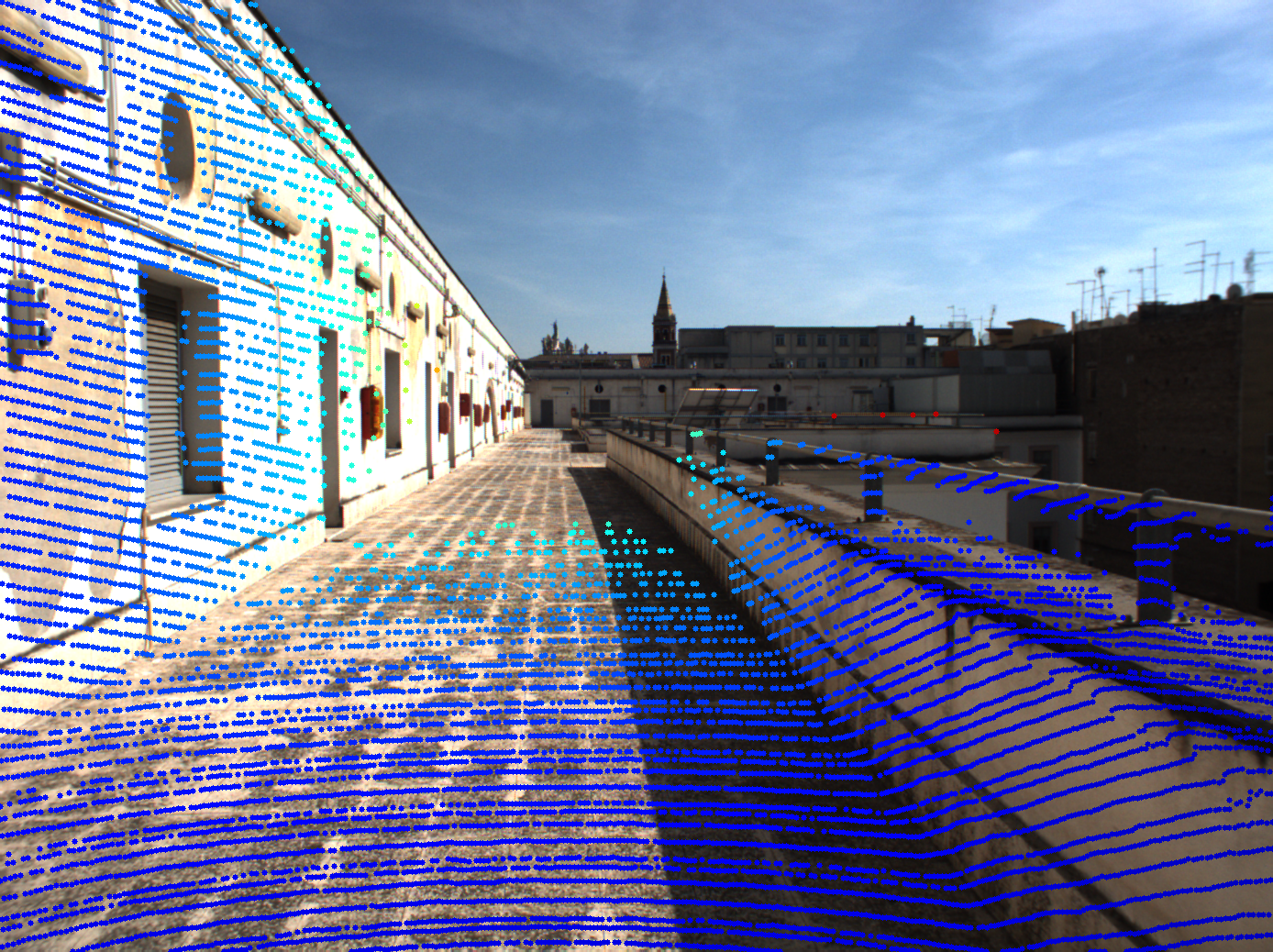}
\end{subfigure}
\begin{subfigure}[b]{0.49\textwidth}
	\centering
	\includegraphics[width=\textwidth]{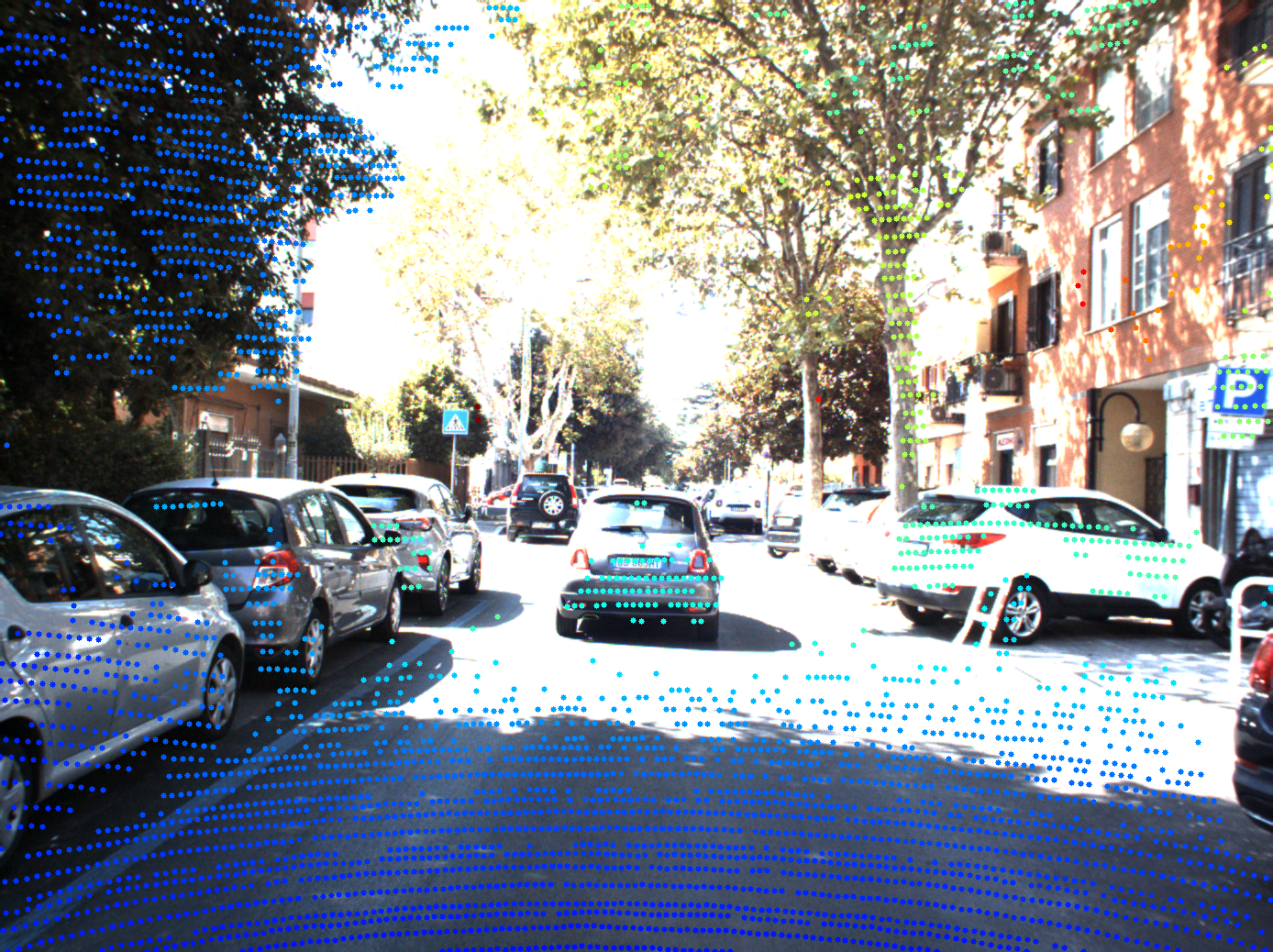}
\end{subfigure}
\begin{subfigure}[b]{0.49\textwidth}
	\centering
	\includegraphics[width=\textwidth]{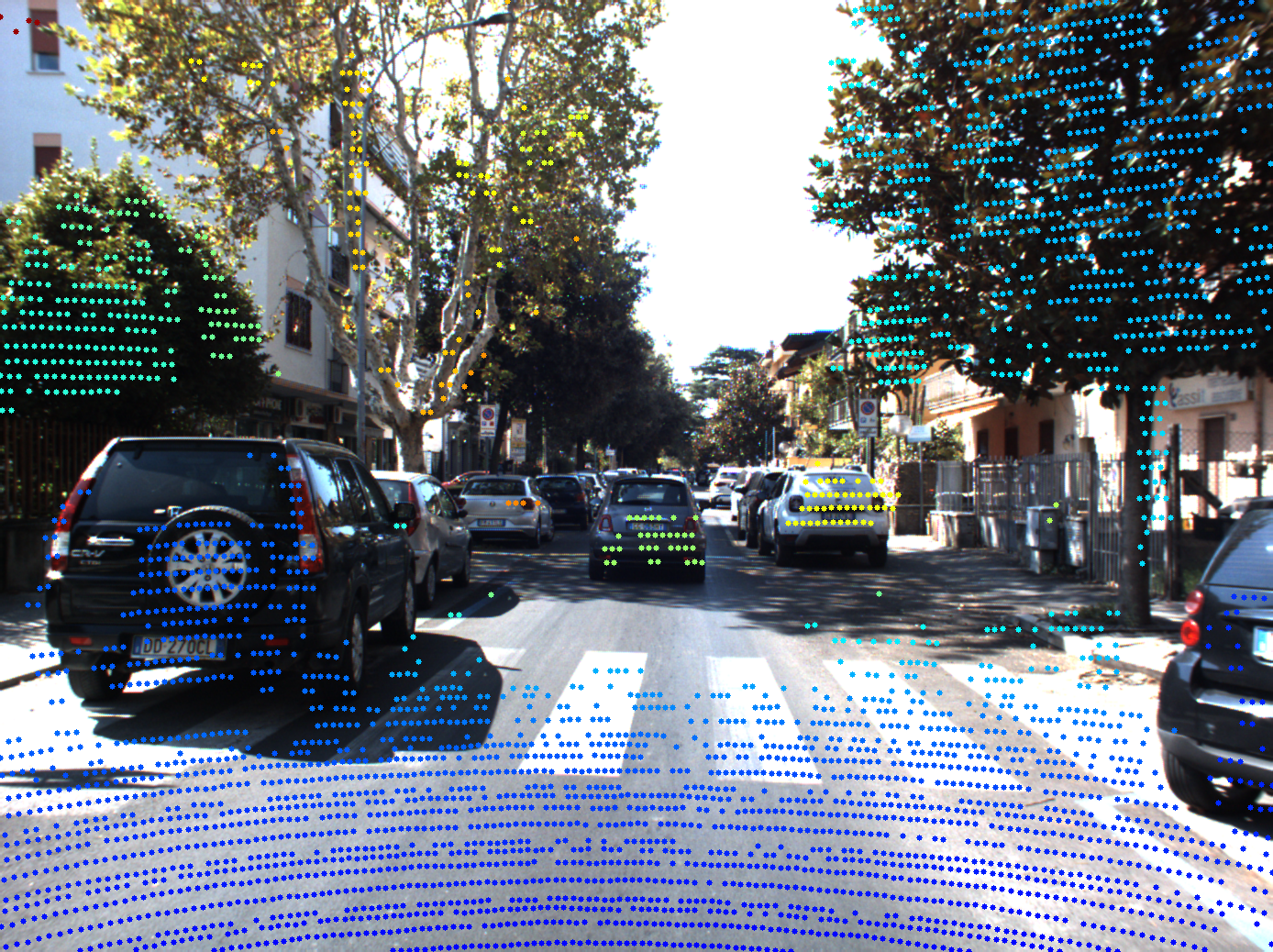}
\end{subfigure}
\begin{subfigure}[b]{0.49\textwidth}
	\centering
	\includegraphics[width=\textwidth]{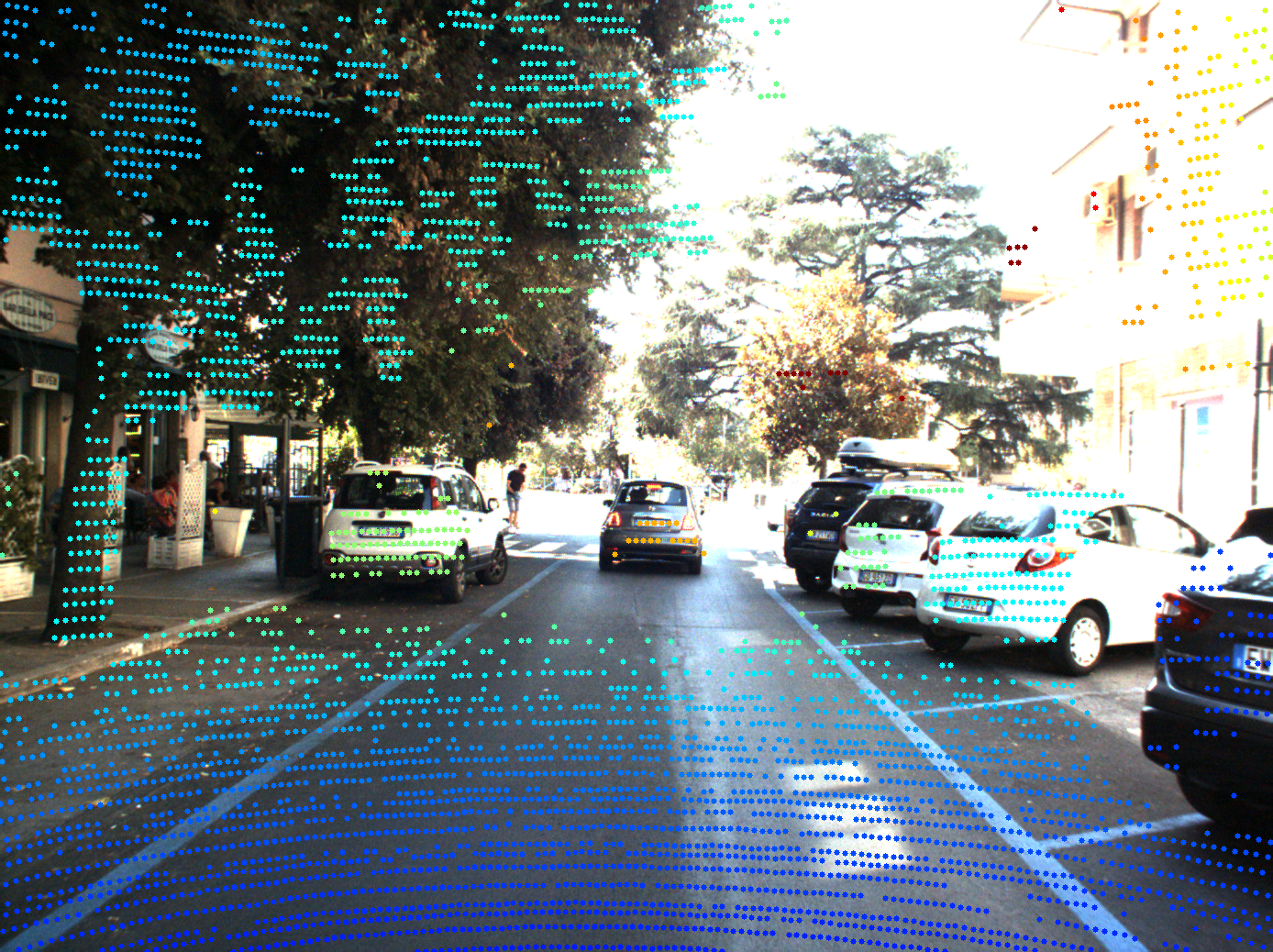}
\end{subfigure}
\caption{Qualitative samples showing LiDAR cloud projection on RGB image.}
\label{fig:qualitative}
\end{figure*}

Moreover, \figref{fig:motivation} and \figref{fig:qualitative} show the reprojection onto the right camera respectively of the fisheye and wide baseline \rgb~sensor. In the latter, the large parallax between the sensors leads to strong occlusions effects, that have been mitigated with an hidden point removal algorithm \cite{katz2007}.

%
%
%
%
%

In summary, our evaluation indicates that our method is capable of generating extrinsic estimates that are comparable or superior to those obtained using other state-of-the-art approaches. It is important to note that careful consideration is required when selecting a minimal number of measurements. However, our experiments clearly demonstrate that the accuracy of these estimates improves as the number of measurements increases.

\section{Conclusion}
\label{sec:conclusion}

In summary, our paper introduces a simple and effective method for accurately estimating extrinsic parameters between \lidars~and \rgb~sensors. By leveraging the inherent planarity of standard calibration patterns, we establish common observations between these sensors, greatly simplifying the calibration procedure.
Our experiments show that planar features mitigate the LiDAR noise, leading to accurate results even with common A3/A4 calibration patterns. Finally, we also release an open source C++ implementation to benefit the community.


%


\bibliographystyle{plain}


\bibliography{robots}
\end{document}

%% file: notation.tex
\usepackage{amsopn}

\newcommand{\bt}{\mathbf{t}}

\newcommand{\bA}{\mathbf{A}}

\newcommand{\bK}{\mathbf{K}}

\newcommand{\bX}{\mathbf{X}}
\newcommand{\bZ}{\mathbf{Z}}
\newcommand{\bR}{\mathbf{R}}

\newcommand{\bbR}{\mathbb{R}}
\newcommand{\bbSE}{\mathbb{SE}}
\newcommand{\bbSO}{\mathbb{SO}}

\newcommand{\bn}{\mathbf{n}}

\newcommand{\bp}{\mathbf{p}}

\newcommand{\bDelta}{\mathbf{\Delta}}

\newcommand{\bDeltaX}{\mathbf{\Delta X}}
\newcommand{\bDeltat}{\mathbf{\Delta t}}
\newcommand{\bDeltaR}{\mathbf{\Delta R}}

\DeclareMathOperator*{\argmin}{argmin}
\DeclareMathOperator*{\atantwo}{atan2}

\def\g2o{$g^2o$}
\def\t2v{\mathrm{log}}
\def\v2t{\mathrm{exp}}
\def\ev2t{\mathrm{ev2t}}

\def\skew#1{{\lfloor{#1}\rfloor}_\times}
